\documentclass[sigconf,screen,authorversion,nonacm]{acmart}

\usepackage{fancyhdr}
\AtBeginDocument{%
    \addtolength{\footskip}{2.0\baselineskip}%
    \fancyfoot[L]{\textit{\textbf{Preprint --- do not distribute.}}}%
}

\usepackage{dsfont}
\usepackage{subcaption}

\usepackage{algorithm}
\usepackage{algpseudocode}
\AtBeginDocument{%
  \providecommand\BibTeX{{%
    \normalfont B\kern-0.5em{\scshape i\kern-0.25em b}\kern-0.8em\TeX}}}

\setcopyright{acmcopyright}
\copyrightyear{2022} 
\acmYear{2022} 
\acmConference[FAccT '22]{2022 ACM Conference on Fairness, Accountability, and Transparency}{June 21--24, 2022}{Seoul, Republic of Korea}
\acmBooktitle{2022 ACM Conference on Fairness, Accountability, and Transparency (FAccT '22), June 21--24, 2022, Seoul, Republic of Korea}
\acmPrice{15.00}
\acmDOI{10.1145/3531146.3533211}
\acmISBN{978-1-4503-9352-2/22/06}

\begin{document}

\title{ABCinML: Anticipatory Bias Correction in Machine Learning Applications}

\author{Abdulaziz A. Almuzaini}
\affiliation{%
  \institution{Department of Computer Science, Rutgers University}
  \streetaddress{}
  \city{New Brunswick, NJ}
  \country{USA}
  }
\email{abdulaziz.almuzaini@rutgers.edu}

\author{Chidansh A. Bhatt}
\affiliation{%
  \institution{Thomas J. Watson Research Center, AI \& Hybrid Cloud, IBM}
  \streetaddress{}
  \city{Yorktown Heights, NY}
  \country{USA}
  }
\email{chidansh.amitkumar.bhatt@ibm.com}

\author{David M. Pennock}
\affiliation{%
  \institution{Department of Computer Science, Rutgers University}
  \streetaddress{}
  \city{New Brunswick, NJ}
  \country{USA}
  }
\email{dpennock@dimacs.rutgers.edu}

\author{Vivek K. Singh}
\affiliation{%
  \institution{School of Communication and Information, Rutgers University}
  \streetaddress{}
  \city{New Brunswick, NJ}
  \country{USA}
  }
\email{v.singh@rutgers.edu}

\renewcommand{\shortauthors}{Almuzaini et al.}

\begin{abstract}
  The idealization of a static machine-learned model, trained once and deployed forever, is not practical. As input distributions change over time, the model will not only lose accuracy, any constraints to reduce bias against a protected class may fail to work as intended. 
  Thus, researchers have begun to explore ways to maintain algorithmic fairness over time. One line of work focuses on \textit{dynamic learning}: retraining after each batch, and the other on \textit{robust learning} which tries to make algorithms robust against all possible future changes. Dynamic learning seeks to reduce biases soon \textit{after} they have occurred and robust learning often yields (overly) conservative models. We propose an anticipatory \textit{dynamic learning} approach for correcting the algorithm to mitigate bias \textit{before} it occurs.  Specifically, we make use of anticipations regarding the relative distributions of population subgroups (e.g., relative ratios of male and female applicants) in the next cycle to identify the right parameters for an importance weighing fairness approach. Results from experiments over multiple real-world datasets suggest that this approach has promise for anticipatory bias correction.
\end{abstract}



\maketitle

\section{Introduction}

    Machine learning (ML) algorithms have been widely used in numerous aspects of our daily lives ranging from simpler tasks such as digit recognition to more complicated and sensitive tasks such as risk assessments \cite{angwin2016machine}, job hiring \cite{miller2015can}, loan lending \cite{fuster2020predictably}, sentiment analysis \cite{kiritchenko2018examining, almuzaini2020balancing},  facial analysis \cite{buolamwini2018gender, alasadi2019toward}, college admission \cite{waters2014grade, santelices2010unfair} and health care \cite{obermeyer2019dissecting, ahmad2020fairness}. Past work has often focused on optimizing the performance (i.e., accuracy) by devising various methods and frameworks, while implicitly ignoring the ethical aspects of the ML models. Recently, the artificial intelligence (AI) research community has started to examine the fairness aspects in the trained state-of-the-art ML applications and significant bias has been detected in various domains. For instance, risk assessment tools were found to discriminate against black people \cite{angwin2016machine}, facial recognition algorithms had higher false positive rate for minority groups including black and hispanic groups \cite{buolamwini2018gender}, and natural language processing (NLP) models, such as hate speech detection and sentiment analysis models were found to be implicitly biased against different groups of the public \cite{almuzaini2020balancing, hutchinson2020social}.  Several in-depth surveys cover these and related findings \cite{berk2021fairness, mitchell2018prediction, makhlouf2020applicability, caton2020fairness, hutchinson201950, mehrabi2019survey, suresh2019framework, friedler2019comparative}.
    
    Although various fairness approaches have been proposed recently in the literature to tackle the issue of bias, the majority of them deal with the conventional static ML training process. Unfortunately, this approach might not generalize well in the future since the world is \textit{non-stationary}, and that could lead to model failures. 
    For instance, a language model that is trained on past data might struggle to generalize well for future data that they have not seen before (e.g., "COVID-19" and "universal lockdown") \cite{lazaridou2021pitfalls}. Also, in scenarios where models learn from future users' inputs (i.e., users behaviors are fed back to a model) without any checks or balances, this could cause the model to learn spurious correlations or even mirror racist or toxic language over time. 

    
    To address the aforementioned problems in both the accuracy and fairness aspects, researchers have proposed various solutions to help ML models generalize well in the future. Currently, there are two main approaches that deal with bias dynamics and early prevention methods: (1) bias detection in a dynamic learning paradigm followed by a correction \cite{iosifidis2019fairness, lazaridou2021pitfalls}, and (2) early prevention using generalization methods such as \textit{robustness} or \textit{domain adaptation} to prepare a model for any issues or shifts that might occur in the future (e.g., ensuring ``worst-case optimal'' results) \cite{koh2021wilds, adragna2020fairness, rezaei2020robust, singh2021fairness}. Although these methods are reasonable and effective, they suffer from some limitations.  In dynamic learning, bias might be corrected \textit{after} a user gets affected which might be harmful for that user and her group. In the case of early prevention via robustness, the model will not have access to future data or retraining, hence these approaches are likely to be either ineffective or overly conservative. 
    

    In scenarios where models interact with the public (e.g., dynamic modeling), it is better to detect bias early and then apply the mitigation methods before it exacerbates. Such techniques (like almost all bias reduction approaches) often do not eliminate bias completely but can significantly reduce its scale, which can be useful in practical settings. In an idealized case, having access to future data will allow the algorithm to tune parameters to obtain desired accuracy and fairness. In practical scenarios, a perfect estimation of future data is impossible. However, there are multiple domains where future data is not completely independent of past data and certain macro-properties of the data follow predictable patterns (e.g., monotonic increase in female representation in college applicants).\footnote{\url{https://www.theatlantic.com/ideas/archive/2021/09/young-men-college-decline-gender-gap-higher-education/620066/}} 
    
    Here we assume that we do not have access to specific instances of future data, but certain macro-properties of the data, for example the relative distributions of population sub-groups (e.g., male, female, wealthy, poor) can be estimated with reasonable accuracy for the next time instance. For instance, while estimating the exact feature description of every new college application for the next year might be very difficult, estimating the relative percentage of applicants from the unprivileged group for the next cycle might be possible. Identifying the relative distributions for the privileged (and unprivileged) groups opens the door for a number of data pre-processing techniques, wherein relative weights or normalizations are undertaken based on the group representations.
    
    Lastly, while there have been multiple metrics for quantifying bias proposed in the past literature \cite{mehrabi2019survey}, most of them have implicitly assumed a static world model. Especially, in ML-fairness literature most of the commonly used metrics (e.g., accuracy and error rate parity) focus on scores derived from the confusion matrix where results across time are collapsed into a single representation. Hence, with the growing interest in temporal aspects of bias, we posit that there is a need to \textit{reify} time in fairness metrics, and newer metrics like temporal stability are critical for evaluating ML models.\footnote{Code: \url{https://github.com/Behavioral-Informatics-Lab/ABCinML}}
    
    Our main contributions in this paper are to:
    \begin{itemize}
        \item add empirical evidence to the literature demonstrating that:{ (a) bias fluctuates frequently and is rarely stable,} and (b) static learning is more likely to not generalize well for both accuracy and fairness, 
        \item propose a framework for dynamic learning along with the anticipation component that can help mitigate bias \textit{before} it happens, and
        \item propose newer metrics for quantifying bias in temporally evolving settings.
    \end{itemize}
    
    The rest of the paper is organized as follows. In Section 2, we provide
    an overview of the related work. Then, Section 3 describes the
    proposed method along with the limitations of the existing baselines. The experimental setup and results are provided in Sections 4 and 5. Finally, in Section 6, a summary and future directions are shared.

\section{Related Work}

    In the last few years, significant research effort has been devoted to fairness in machine learning and as a result, several definitions, metrics and methods have been proposed to address both bias detection and mitigation aspects \cite{caton2020fairness, makhlouf2020applicability, mehrabi2019survey, hovy2021five}. These measurements might be domain specific and unfortunately have limitations. Several authors provide mathematical proofs of the impossibility of simultaneously satisfying different proposed metrics \cite{berk2021fairness, kleinberg2016inherent, chouldechova2017fair, friedler2016possibility}. 
    
    Following bias measurements, researchers proposed various mitigation methods which include: \textit{pre-processing} in which the dataset has to be corrected (i.e., modified to support fairness) prior to the modeling~\cite{kamiran2012data}, \textit{in-processing} where a model itself is being corrected by using constrained optimization or penalty methods~\cite{zafar2017fairness, bechavod2017penalizing}, and \textit{post-processing} where the model's prediction distributions are modified in order for the model to be fair~\cite{hardt2016equality}. 
    
    Addressing the limitations of static learning, recently Lazaridou et al.~\cite{lazaridou2021pitfalls} examined the performance of a language model performance when a model is trained on past data and used to generalize to future data. The model performs worse on data that are far away from the training periods which leads to a failure in the temporal generalization aspect. Proposed solutions to mitigate this kind of issue are by retraining or adapting the model repeatedly.

    Applying generalization methods by training a model that is robust to various distribution shifts, researchers examine the effects of robustness on fairness. Iosifidis et al.~\cite{iosifidis2019fairness, iosifidis2020mathsf} apply pre-processing and distribution shifts methods in streaming classification framework in which the dataset or the model has to be corrected at each time step so the model stays stable. Singh et al.~\cite{singh2021fairness} utilize causal learning to build a model that is insensitive to distribution shifts that might occur in the features (i.e., covariate shift). Rezaei et al.~\cite{rezaei2020robust} try to mitigate bias under covariate shift as well by using pre-processing and in-processing methods simultaneously to build a robust and a fair model. Lastly, applying invariant risk minimization (IRM) instead of the conventional empirical risk minimization (ERM) method forces the model to learn features that are invariant to any distribution shifts (i.e., forcing the model to not rely on spurious relationships)~\cite{adragna2020fairness}.
    

\section{Proposed Method}

    Our approach includes an anticipation component to better address fairness in dynamic learning. In dynamic learning, unlike static learning, the model keeps updating its knowledge as new data arrives. We assume there are two disjoint batches that have data from different sets such as: the current and the future. A model learns during the current period and once the future period becomes available, a model uses it to measure its performance. If the performance shows some sort of generalization issue, the model needs to be updated. This kind of issue is referred to as distribution shift \cite{storkey2009training, lazaridou2021pitfalls}.
    
    Distribution shifts can be divided into multiple classes, namely concept, features or temporal shift \cite{storkey2009training}. In our formulation, we assume a temporal shift exists in the data and this type of shift is what causing the model to be unfair. We examine a specific example of temporal shift which is \emph{selection bias} where different batches of the data have different ratios of both the class labels and the sensitive attributes. Thus a model trained with such data might learn well with the majority groups and not with the minority groups. This imbalance is likely going to lead to the issue of bias. Specifically, we assume (and empirically validate) that we are able to estimate the relative distributions of the class labels and the sensitive attributes at an accuracy level that is good enough to support bias correction.  
    
    To mitigate this issue of \emph{selection bias}, we want to encourage the model to learn from both groups equally regardless of the data imbalance. To do so, we adapt a \emph{pre-processing} method developed by \cite{kamiran2012data} which is a \emph{reweighing} method that gives weights to different groups based on their representations in the dataset with the class labels in order to force the model to learn fairly. Different from \cite{kamiran2012data}, we not only correct the dataset based on the current representations but also apply a correction based on the relative distributions that we expect in the future.

 \begin{algorithm*}
        \caption{Anticipatory Bias Correction}\label{}
        \begin{algorithmic}[1]
            \Procedure{ABC}{$\; B_{1:t}(A, Y),\; S , \; \alpha$}    \Comment{data until time $t$, S: window length, $\alpha$: smoothing factor}
            
            \State$P_{\tilde{B}_{t+1}}(A, Y)\gets \dfrac{P_{B_{t}}(A, Y) +  P_{B_{t-1}}(A, Y) + ... + P_{B_{t-S}}(A, Y) }{S}$ \Comment{forecast the relative distributions for batch $\tilde{B}_{t+1}$}
            
            \State$ W_{B_{t}}(A, Y) \gets Reweighing$     \Comment{use Eq. (1) to get weights for batch $B_{t}$}
            
            \State$ W_{\tilde{B}_{t+1}}(A, Y) \gets Reweighing$ \Comment{use Eq. (1) to get weights for batch $\tilde{B}_{t+1}$}
            
            \State$ W_{New} \gets \alpha \times W_{B_{t}}(A, Y) + ( 1 - \alpha) \times W_{\tilde{B}_{t+1}}(A, Y) $       \Comment{acquire the new weight}
            
            \State \textbf{return} $f_t(B_{t}, W_{New})$ \Comment{learn a new classifier with the new weight}
            \EndProcedure
        \end{algorithmic}
    \end{algorithm*}

\subsection{Preliminaries}

    We define a random variable $V$, representing the input variables (i.e., covariates). We can divide $V$ into $(A, X)$, where $A$ is a binary random variable representing the \textit{sensitive} attribute and $X$ represents the \textit{non-sensitive} attributes. Also, we represent the ground truth by a binary random variable $Y$. Each instance $v \in V$ has a label $y \in Y $ and a sensitive attribute $a \in A$ such that: $y = \{y^{+}, y^{-}\}$, where \textit{+, (-)} represents \textit{positive class (negative class, respectively)}, and $a = \{a^{+}, a^{-}\}$ where \textit{+, (-)} represents \textit{privileged (unprivileged, respectively)}. We utilize a function $f \colon V \to \{y^{+}, y^{-}\}$, representing a binary classifier. 
    
    We assume that the data arrive sequentially in batches $\{B_1, B_2, ...,\}$ in which each batch $B_t$ has a collection of $j$ instances drawn from $V$ such as $B_t = (v_1^t, v_2^t, ..., v_j^t)$ and $t$ represents the time dimension. Following the dynamic learning settings, we have disjoint sets representing the ($B_{t}$) and ($B_{t+1}$) in which we refer to as \textit{current} and \textit{future} batches, respectively. Specifically, $B_{t}$ represents the current data we train $f$ on, (i.e., $f_{t}(B_{t}) $), whereas $B_{t+1}$ is the future data that we use to evaluate $f_{t}$, (i.e., $f_{t}(B_{t+1})$). Since $B_{t}$ and $B_{t+1}$ are disjoint, we assume the distribution is \textit{non-stationary}, i.e., $P_{B_{t}}(V, Y) \neq  P_{B_{t+1}}(V, Y)$.

\subsection{Pre-Processing Method}

    To mitigate the \textit{discrimination} in the dataset, we adapt a popular pre-processing method to reduce the discrimination in the dataset before applying the learning model \cite{kamiran2012data}. The \textit{Reweighing} method assigns different weights \textit{w} for each sub-populations with regards to their representations in the dataset. Hence, positive outcome instances for the unprivileged group should be valorized, while negative outcome instances for the unprivileged group can be given lower weights. Specifically, $P(A= a^{-}, Y= y^{+})$ will have higher \textit{weights} compared to $P(A= a^{+}, Y= y^{+})$, whereas $P(A= a^{-}, Y= y^{-})$ will have lower \textit{weight} compared to $P(A= a^{+}, Y= y^{-})$. Therefore, for each $B_{t}$, the weights assigned as follows:
    
    \begin{equation}
        W_{B_{t}} (A, Y) = \frac{P_{expected}(A, Y)}{P_{observed}(A, Y)}
    \end{equation}
    
    where $P_{expected}(A, Y)$ can be estimated from the dataset as the following:
    
    \begin{align*}
        P_{expected}(A, Y) &= P(A) \times P(Y) \\
        &= \frac{|\{ A = a \}|}{|B_{t}|} \times \frac{|\{ Y = y \}|}{| B_{t}|}
    \end{align*}
    
    and $P_{observed}(A, Y)$ would be:
    \begin{align*}
        P_{observed}(A, Y) &= \frac{| \{A = a, Y = y\}|}{|B_{t}|}
    \end{align*}
    
    Therefore, we could use any learning models which permits applying these weights in their frameworks.  
    
\subsection{Models}

    In this section, we provide details of the modeling techniques used in the experiment:
    
    \textbf{ 0) Vanilla setting: Train once for accuracy, Don't mitigate bias, Test sequentially.} In this baseline, we don't address fairness at all and we want to examine the behavior of bias through time (i.e., whether it is stable or fluctuating). While simple, this is the most common setting used in current machine learning implementations. 
    
    \textbf{ 1) Static setting: Train once for accuracy and bias mitigation, Test sequentially.} We simulate the typical static learning where a model is only trained and corrected once on a static dataset then deployed. Thus, a model might be initially \textit{discrimination-free} but fails in the future due to changes in the underlying distribution.
    
    At the beginning of the training,  we assume $B_{t}$ arrives with the corresponding features and labels (i.e., $V$ and $Y$, respectively) and to \textit{mitigate} bias before training, we apply the \textit{reweighing} method on this batch to get $W_{B_{t}}(A, Y)$. Using $B_{t}$ and its corresponding weight $W_{B_{t}}$, we train a classifier $f_{t}(B_{t},W_{B_{t}})$. Then, we use $f_{t}$ to evaluate incoming batches $\{t+1, t+2,..., n\}$.

    \textbf{ 2) Dynamic setting: Train for accuracy and bias mitigation sequentially, Test sequentially.} Addressing the limitations of baselines (0) and (1), we overcome this issue of training once by re-training the model \textit{continuously} every time a batch arrives. By doing so, we keep the model up-to-date and as a result, we reduce the effects of the temporal and the distribution shift. 
    

    \textbf{ 3) Anticipatory Bias Correction (ABC).} We propose an anticipatory model that  utilizes future estimates of the upcoming batches. In a variety of applications, especially when a dataset doesn't follow the i.i.d assumptions, the underlying distribution might show some behaviors that might be forecasted. In this work, we utilize a basic yet effective forecasting model to anticipate some \textit{macro-properties} about the future batches. Specifically, we use a \textit{Moving Average} model to \textit{anticipate} the incoming batch's relative distributions in $\tilde{B}_{t+1}$, i.e., $P_{\tilde{B}_{t+1}}(A=a, Y=y)$ as following:
    
    \begin{equation}
    \small
    P_{\tilde{B}_{t+1}}(A, Y) = \frac{P_{B_{t}}(A, Y) +  P_{B_{t-1}}(A, Y) + ... + P_{B_{t-S}}(A, Y) }{S}
    \end{equation}
    
    where $S$ represents the window's length and $\tilde{B}_{t+1}(A, Y)$ is the estimated distribution for batch $t+1$. Note that we do not assume that the actual data points from $B_{t+1}$ are available or estimated. Rather, we hypothesize that the data points from $B_{t}$, when weighed according to the anticipated ratios for $B_{t+1}$ would already be useful in mitigating the bias levels in $B_{t+1}$. Using Eq.(1) and Eq.(2), we will be able not to only mitigate bias for the current time step but also for the future as well. Doing so, will help us \textit{prevent} bias before it shows up in the output of the algorithm. (See Algorithm 1).
    
    To apply the reweighing method, at time $t$: we will have $W_{B_{t}}(A, Y)$ and $W_{\tilde{B}_{t+1}}(A, Y)$, weight estimates for the current and the future data, respectively. We combine these \textit{weights} to have a new weight $W_{New}$ as the following:
    
    \begin{equation}
        W_{New} = \alpha \times W_{B_{t}}(A, Y) + ( 1 - \alpha) \times W_{\tilde{B}_{t+1}}(A, Y) 
    \end{equation}
    
    where $\alpha \in [0, 1]$. We apply this weighted approach to balance the weights between current data and the future data. Lower $\alpha$ will put much emphasis on the future data, whereas higher $\alpha$ focuses on the current data. Lastly, we build a classifier using the the new weight, i.e., $f_t(B_{t}, W_{New})$. 

\subsection{Models Assessments}

To assess the model discrimination, we utilize three different bias metrics that are commonly used in the fairness literature \cite{makhlouf2020applicability}. However, as these metrics build upon confusion matrices that collapse variations over time into a single representation, we complement these traditional snapshot metrics with some newer temporal metrics. 

\subsubsection{Snapshot metrics} \hfill\

\textbf{\small{Statistical Parity Difference ($\Delta$ S.P)}} measures the disparity of being assigned to a positive class for individuals from different groups. In other words, a fair model requires the predictions to be statistically independent from the sensitive attributes:

\begin{equation}
    \small
    \Delta \; S.P = | P(\hat{Y} = 1 | A = 0) - P(\hat{Y} = 1 | A = 1)|
\end{equation}

\textbf{\small{Equal Opportunity ($\Delta$ TPR)}} measures the disparity of the True Positive Rate (TPR) for individuals from different groups. In other words, a fair model requires equal TPR for individuals from different groups:

\begin{equation}
    \small
    \Delta  \; TPR = | P(\hat{Y} = 1 | Y = 1, A = 0) - P(\hat{Y} = 1 | Y = 1,  A = 1) |
\end{equation}

\textbf{\small{Predictive Equality ($\Delta$ FPR)}} measures the disparity of the False Positive Rate (FPR) for individuals from different groups. In other words, a fair model requires equal FPR for individuals from different groups:

\begin{equation}
    \small
    \Delta  \; FPR = | P(\hat{Y} = 1 | Y = 0, A = 0) - P(\hat{Y} = 1 | Y = 0,  A = 1) |
\end{equation}

All the above equations measure the absolute difference where a lower value indicates a fair model and a larger value indicates a biased model. 

For the model accuracy, we use the \textbf{\small{Area Under the ROC Curve (AUC)}} since it is robust to class imbalance. Often, it is impossible to achieve a low value for all the fairness measures at the same time since different metrics are domain-specific and have potentially contradictory assumptions \cite{friedler2016possibility}.   

\subsubsection{Temporal Metrics} \hfill\

    To allow for a richer understanding of bias in temporally evolving settings, we propose some temporal metrics to examine bias. Following related literature in fairness in ML, time series analysis, and capturing the performance of dynamical systems \cite{g2020methods,gupta2020too, teinemaa2018temporal}, we posit that the metrics should be able to capture the following aspects:  
    \begin{itemize}
    \item{the worst case performance of the system}
    \item{the fluctuations in the performance of the system}
    \end{itemize}

    These metrics can be helpful in capturing the underlying behaviors in which bias rate can change, and thus can be used for evaluation and monitoring purposes. Just like there exist multiple metrics to measure aspects related to accuracy (e.g., AUC, true positive rate, precision, recall, f-measure), we expect a subset of these temporal bias metrics to be used in a given scenario based on the application priorities. Following substantive existing literature in the space, we use $\Delta$ (i.e., a disparity treatment between two sensitive groups) as the starting point to quantify bias. This $\Delta$ can be operationalized over any of the metrics that capture performance (e.g., accuracy, TPR, statistical parity etc.) as appropriate in the given context.
    
    \paragraph{Worst case performance} \hfill
    
    The bias level of the system may fluctuate over time, hence in many scenarios it is important to understand the worst case bias level.
    
    \textbf{\small{Maximum Bias (MB)}} is defined as the maximum bias observed across all time steps $N$:
    \begin{equation}
        MB = \max\limits_{1\leq i\leq N} \Delta_{i}
    \end{equation}
    
    \textbf{\small {MB}} quantifies the most biased performance that can be expected from the system if we cannot control the time at which the user will interact with the system. A stable and a fair model is expected to have a low MB.

    \paragraph{Fluctuation in performance} \hfill
    
    An ideal ML system will have low bias and that level of bias will not change dramatically over time. Dramatic changes over time will yield very different performances to different users who may happen to use the system on neighboring time points.

    \textbf{\small Temporal Stability (TS)} is defined as the average absolute deviation of consecutive bias windows:  
    \begin{equation}
        TS = \frac{1}{N} \sum_{i=2}^{N} |\Delta_{i} - \Delta_{i-1}|
    \end{equation}
    
    Since we are using a dynamic learning approach, consecutive time steps are expected to have yield similar performance. Therefore, we propose a metric the examine the \emph{adjacent/local fluctuation} with respect to the previous time steps. If the model were to be stable and fair, we expect a lower estimate.
    
    \textbf{\small Maximum Bias Difference (MBD)} is defined as the maximum bias difference between consecutive bias windows: 
    \begin{equation}
        MBD = \max\limits_{2\leq i\leq N} |\Delta_{i} - \Delta_{i-1}|
    \end{equation}
    
    This metric is examining the \emph{sudden change} that might happen in the bias rate during modeling. A larger sudden change may suggest a phase transition or mishap that needs to be looked into. A \textit{stable} fair model will persistently generate a lower change over consecutive time windows and hence will have a lower MBD.  

\begin{table}[h]
  \centering
  \caption{Datasets overview}
    \begin{tabular}{l*{6}{c}r}
    \textbf{Dataset}              & \textbf{Period} & \textbf{No. Samples} & \textbf{Sensitive} & \textbf{Target} \\
    \hline
    Funding      & 2005-2013 & 612,262 & Poverty-Level & Funded?    \\
    Toxicity     & 2015-2017 & 60,287 & Gender & Toxic?     \\
    Adult       & 2014-2018 & 636,625 & Race & >=50K?   \\
   
    \end{tabular}
    
\vspace*{10pt}\end{table}

\section{Experiments}
    Our approach is  applicable to datasets that have a temporal dimension in order to test the temporal shift and generalization problems. Unfortunately, most of the current fairness benchmarking datasets do not have the temporal aspect, i.e., the data does not come with explicit timestamp. However, there are three temporal datasets that have been used recently for fairness applications and we focus on these for this work. 

\subsection{Datasets}

    We validate our approach with the following datasets that have a temporal dimension.
    
    \textbf{Funding} is a dataset provided by DonorsChoose\footnote{\url{https://www.donorschoose.org/}}, an organization that facilitates educational projects funding posted by teachers in the United States and encourages local community members to support teachers' projects by donation. The 2014 Data Mining and Knowledge Discovery competition (KDD) publicly released the dataset to encourage ML practitioners to build a predictive model to find which projects are more likely to be funded in the future based on information/attributes related to the projects, schools, teachers and funding \cite{KDD, lamba2021empirical}. To address the fairness aspect, we utilize the "poverty-level" feature as a sensitive attribute in order to see if there is any discrimination with respect to the wealth level of the school district. We use data records from 2005-2013 and employ monthly temporal modeling. We utilize a window of 3 for forecasting (e.g., using the first three months to forecast some knowledge about the $4^{th}$ month).

    \textbf{Civil Comments Toxicity}\footnote{\url{https://www.kaggle.com/c/jigsaw-unintended-bias-in-toxicity-classification}} is a corpus of comments collected for 2 years (2015-2017) in order to address bias in the toxicity classification applications. The dataset has been used to address the spurious correlations between the sensitive attribute and the probability of toxicity. The prediction task is to predict if a comment is toxic or not. To measure the fairness aspect, the task is to measure bias in comments in which an identity or an ethnicity has been mentioned. In this task, we focus on the gender bias \cite{borkan2019nuanced, adragna2020fairness}.  We  experiment with a monthly temporal modeling in which a window of 5 has been used. 
    
    \textbf{Adult} is an extension of the popular 1994 adult dataset that have been widely used in the ML-fairness literature \cite{kohavi1996scaling}. We used a newly released version in which the dataset is collected between 2014-2018 and span across US states \cite{ding2021retiring}. The prediction model is to classify whether a person's income will be more than 50K based on the demographic attributes \cite{quy2021survey}. We only utilize a subset of the dataset from the state of California and we focus on the racial bias (i.e., "white" v.s. "black"). For this dataset, we apply a yearly temporal modeling since the monthly measurement is not available. Because of the small time-range, we use a window of 2 for forecasting.  
    
    The datasets' information is summarized in Table 1. 

\section{Results And Discussion}

    Since, in this experiment we are utilizing dynamic learning as apposed to the traditional static learning, we use a re-training approach. Having large data size might allow the model to learn features adequately by being exposed to a variety of data points. Therefore, we train on the current batch and we test on the next. We employ a growing window approach for the training dataset to maximize the learning opportunities for the model. 
    
    To evaluate the model, we restrict the learning hypothesis domain to be a simple classifier, namely "Logistic Regression". For each dataset, we use feature representations suitable for each applications. Specifically, for the Funding dataset, we preprocess the data and apply feature engineering methodology to keep only useful features as suggested by \cite{Sofia2020} yielding 113 features. In the case of the Toxicity dataset, we use a pre-trained word embedding (i.e., Word2vec-100d) to represent linguistic features \cite{Mikolov2013EfficientEO}. Lastly, we use all the 12 features provided by the authors for the Adult dataset \cite{ding2021retiring}.

    \subsection{Temporal Variation}
    \begin{figure*}[ht]
        \centering
         \includegraphics[width=1\linewidth]{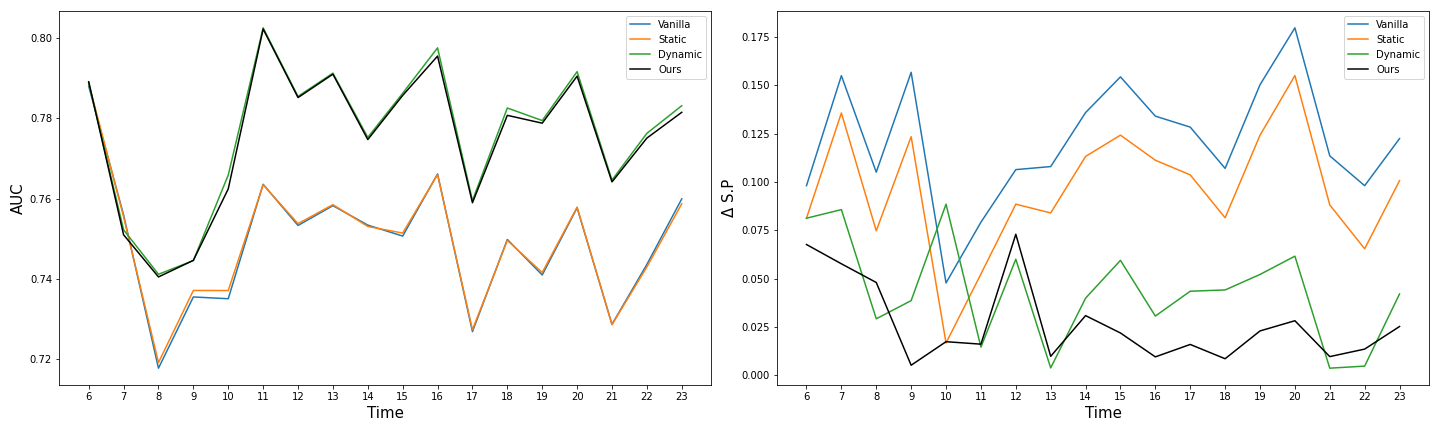}\label{fig:Jigsaw_AUC_SP}
        \caption{Changes in accuracy (AUC) and fairness ($\Delta$ S.P) over time for the toxicity dataset}
        \Description{Changes in accuracy (AUC) and fairness ($\Delta$ S.P) over time for the toxicity dataset}
\vspace*{20pt}\end{figure*}
    
    To examine the temporal dynamics for both the accuracy and bias, we plot the performance of the abovementioned approaches over time for all datasets. Figure 1 shows the baselines' results along with our proposed method for accuracy (AUC) and the primary fairness metric considered ($\Delta$ S.P) in the Toxicity dataset. (Results for other datasets and using other fairness metrics are presented in the Supplementary Material).
    As can be seen, the accuracy of the model (AUC) changed noticeably over time. For baselines 0 (vanilla) and 1 (static), which are the most common approaches for ML implementations, the AUC varied between 0.79 and 0.72. A growing window and the dynamic learning approach (baseline 2) yielded a higher accuracy and that performance was virtually matched by the proposed approach. There is also noticeable fluctuation in bias levels for baseline 0 (range from 0.175 to 0.050). In effect, the results add empirical evidence to the literature demonstrating that: (a) bias is not a static entity and fluctuates frequently, and (b) vanilla and static learning (baselines 0,1) are not likely to generalize well for both accuracy and fairness \cite{ding2021retiring, iosifidis2021online, iosifidis2019fairness, d2020fairness}. 
    
    Further, we notice that baseline 2 performs better (i.e., provides lower bias levels and higher accuracy levels) than baselines 0 and 1. The proposed approach yields lower bias level compared to baseline 2 while maintaining accuracy at levels comparable to baseline 2.

    \subsection{Impact of Future Estimation}
    
    A key question in this work is to study the impact of future estimation on the accuracy and fairness levels of the algorithms. For ease of interpretation and comparison with existing work, we first quantify fairness based on popular (snapshot) metrics and then discuss the impact on proposed temporal metrics in the next subsection.
    
    \begin{table*}[h]
     \caption{Results for performance of different baselines and the proposed approach in different applications. Results reported for accuracy (AUC) and popular `snapshot' fairness metrics.}
    \resizebox{\textwidth}{!}
    {%
    \begin{tabular}{lcccl|cccl|cccl}
    \hline
    \multicolumn{1}{l}{} & \multicolumn{4}{c}{\textbf{Funding}}                              & \multicolumn{4}{c}{\textbf{Toxicity}}                             & \multicolumn{4}{c}{\textbf{Adult}}                                \\ \hline
    \multicolumn{1}{l}{} & AUC ↑          &$\Delta$ S.P ↓         &$\Delta$ TPR ↓          &$\Delta$ FPR ↓          & AUC ↑          &$\Delta$ S.P ↓         &$\Delta$ TPR ↓          &$\Delta$ FPR ↓          & AUC ↑          &$\Delta$ S.P ↓         &$\Delta$ TPR ↓          &$\Delta$ FPR ↓          \\
    \textbf{Vanilla}           & 0.651          & 0.309          & 0.318          & 0.248          & 0.749          & 0.121          & 0.102          & 0.115          & 0.824          & 0.107          & 0.095          & 0.038          \\
    \textbf{Static}           & 0.655          & 0.153          & 0.152          & 0.108          & 0.749          & 0.096          & 0.071          & 0.090          & 0.818          & 0.079          & 0.082          & 0.008          \\
    \textbf{Dynamic}           & \textbf{0.716} & 0.071          & 0.076          & 0.052          & \textbf{0.776} & 0.043          & \textbf{0.044} & 0.038          & 0.822          & 0.074          & 0.076          & \textbf{0.006} \\
    \textbf{Ours}        & 0.714          & \textbf{0.064} & \textbf{0.071} & \textbf{0.049} & 0.775          & \textbf{0.027} & 0.054          & \textbf{0.024} & \textbf{0.826} & \textbf{0.058} & \textbf{0.060} & 0.024          \\ \hline
    \end{tabular}%
    }
    \label{tab:my-table}
\vspace*{30pt}    \end{table*}

    \begin{figure*}[h]
    \begin{subfigure}{\linewidth}
     \begin{center}
      \includegraphics[width=0.9\linewidth]{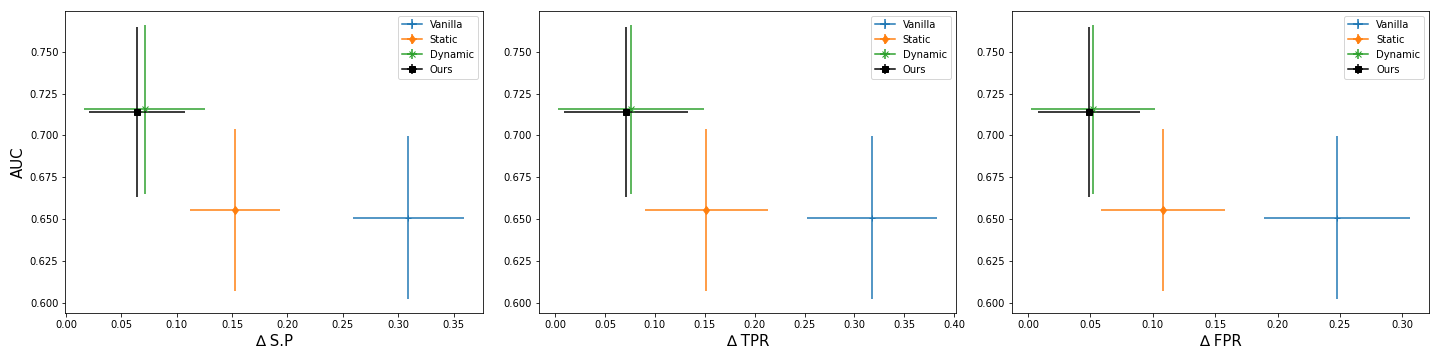}
     \end{center}
      \caption{Funding Dataset}
      \end{subfigure}\par\medskip
      
      \begin{subfigure}{\linewidth}
     \begin{center}
      \includegraphics[width=0.9\linewidth]{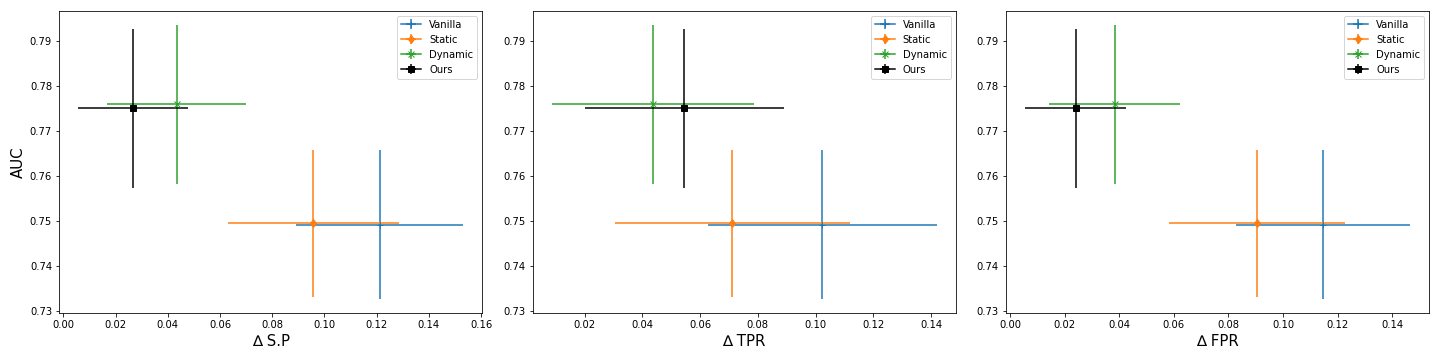}
     \end{center}
      \caption{Toxicity Dataset}
      \end{subfigure}\par\medskip
      \begin{subfigure}{\linewidth}
     \begin{center}
      \includegraphics[width=0.9\linewidth]{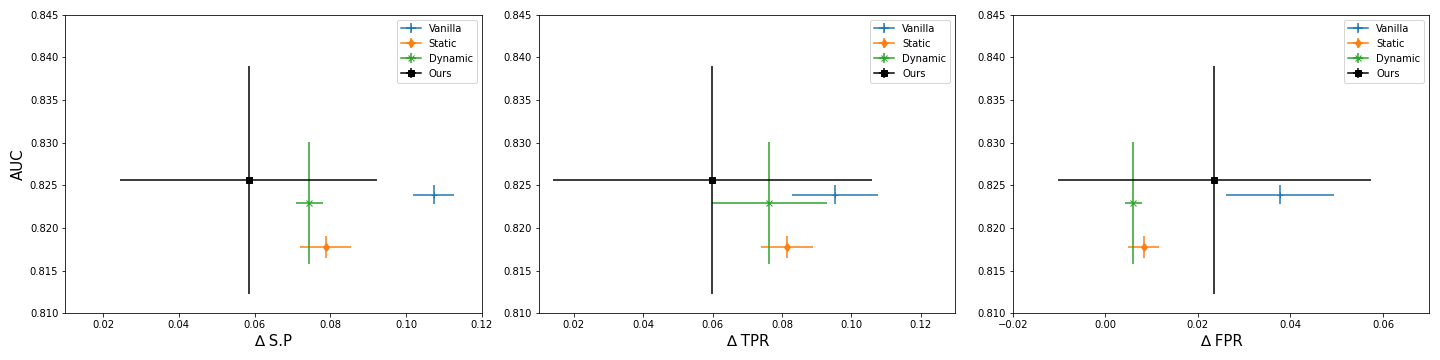}
     \end{center}
      \caption{Adult Dataset}
      \end{subfigure}\par\medskip
      \caption{Results of experimental evaluation for different datasets. Y-axis shows accuracy (AUC) and the X-axis shows a bias metric. Average scores for different approaches are shown as points with with standard deviation shown as a bar. Best models are those that lie in the top left portion of the figure.}
      \Description{Results of experimental evaluation for different datasets. Y-axis shows accuracy (AUC) and the X-axis shows a bias metric. Average scores for different approaches are shown as points with with standard deviation shown as a bar. Best models are those that lie in the top left portion of the figure.}
      \label{fig:FairnessAcc_3datasets}
     
    \end{figure*}

    Table 2 and Figure 2 show the results of the three baselines and the proposed approach. (Note that the $\alpha$ for the proposed approach is chosen for its best performance on the $\Delta$ S.P). As can be seen, the baselines 0 and 1 struggle to mitigate bias \textit{on average}. Baseline 2 is helpful in decreasing the bias discrimination across all the considered metrics with a noticeable reduction. The proposed approach yields the lowest value of bias with regard to $\Delta$ S.P across all datasets (Fig. 2 - left plots). Conceptually, this can be interpreted as importance weighing approach trying to mimic the sampling procedures by having equal representations of different groups with regard to the class labels. Additionally, the proposed approach has successfully reduced the bias across all measurement in the Funding dataset (Fig. \ref{fig:FairnessAcc_3datasets} (a)) and for two out of the three metrics in the other datasets. As suggested by prior literature, different bias metrics may not always be reduced in the same settings. 
    
    We report the impact of $\alpha$ (i.e., relative importance given to current data or the future estimation (Eq. 3)) in Table 3. The value of $\alpha$ ranged from 0 to 1 in which lower value means that the model is more focused on mitigating the bias using the future estimates, whereas a higher value is the opposite. Traditional systems engineering approaches such as Kalman Filtering suggest that different applications and contexts would require different level of importance to be given to the learned parameters from past or current data and the estimates of future data \cite{singh2008coopetitive}. 
    Here we also found different applications have different behaviors. In case of the Funding and Adult datasets, the models perform better when focusing more heavily on the current data with slight reliance on the future estimates  ($\alpha$=0.9). In contrast, in applications that are more prone to distribution shift, such as Toxicity Classification, the model performs better when giving more importance to future estimates ($\alpha$=0.0) \cite{adragna2020fairness}.

    \begin{table*}[h]
    \caption{Results for different values of $\alpha$ and its effects on our approach. Lower value of $\alpha$ focuses on future, whereas higher value focuses on the current.}
    \resizebox{\textwidth}{!}{%
    \begin{tabular}{lcccl|cccl|cccl}
    \hline
                 & \multicolumn{4}{c}{\textbf{Funding}}                              & \multicolumn{4}{c}{\textbf{Toxicity}}                             & \multicolumn{4}{c}{\textbf{Adult}}                                \\ \hline
                 & AUC ↑          &$\Delta$ S.P ↓          &$\Delta$ TPR ↓          &$\Delta$ FPR ↓          & AUC ↑          &$\Delta$ S.P ↓          &$\Delta$ TPR ↓          &$\Delta$ FPR ↓          & AUC ↑          &$\Delta$ S.P ↓          &$\Delta$ TPR ↓          &$\Delta$ FPR ↓          \\
    \textbf{0.0} & 0.713          & 0.107          & 0.115          & 0.084          & \emph{0.775}          & \emph{\textbf{0.027}} & \emph{0.054}          & \emph{\textbf{0.024}} & 0.820          & 0.088          & 0.088          & 0.012          \\
    \textbf{0.1} & 0.714          & 0.101          & 0.108          & 0.077          & 0.775          & 0.028          & 0.051          & 0.025          & 0.823          & 0.094          & 0.095          & 0.018          \\
    \textbf{0.2} & 0.713          & 0.095          & 0.104          & 0.074          & 0.775          & 0.030          & 0.048          & 0.027          & 0.820          & 0.087          & 0.083          & 0.013          \\
    \textbf{0.3} & 0.714          & 0.088          & 0.097          & 0.069          & 0.775          & 0.031          & 0.045          & 0.028          & 0.817          & 0.078          & 0.080          & 0.008          \\
    \textbf{0.4} & 0.714          & 0.083          & 0.092          & 0.063          & 0.775          & 0.032          & 0.044          & 0.030          & 0.822          & 0.089          & 0.090          & 0.013          \\
    \textbf{0.5} & 0.715          & 0.079          & 0.088          & 0.063          & \textbf{0.776} & 0.034          & 0.041          & 0.032          & 0.821          & 0.093          & 0.0901         & 0.023          \\
    \textbf{0.6} & 0.714          & 0.076          & 0.084          & 0.059          & \textbf{0.776} & 0.036          & 0.042          & 0.032          & \textbf{0.825} & 0.101          & 0.098          & 0.026          \\
    \textbf{0.7} & 0.714          & 0.072          & 0.081          & 0.056          & \textbf{0.776} & 0.037          & 0.042          & 0.034          & 0.822          & 0.073          & 0.071          & 0.008          \\
    \textbf{0.8} & 0.714          & 0.067          & 0.074          & 0.054          & \textbf{0.776} & 0.039          & \textbf{0.040} & 0.035          & 0.822          & 0.092          & 0.092          & 0.015          \\
    \textbf{0.9} & \emph{0.714}          & \emph{\textbf{0.064}} & \emph{\textbf{0.071}} & \emph{\textbf{0.049}} & \textbf{0.776} & 0.042          & 0.042          & 0.037          & \emph{\textbf{0.825}} & \emph{\textbf{0.058}} & \emph{\textbf{0.059}} & \emph{0.023}          \\
    \textbf{1.0} & \textbf{0.716} & 0.071          & 0.076          & 0.052          & \textbf{0.776} & 0.043          & 0.044          & 0.038          & 0.823          & 0.074          & 0.076          & \textbf{0.006} \\ \hline
    \end{tabular}%
    }
    \label{tab:my-table}
    \end{table*}
    
    \subsection{Temporal Metrics Evaluation}
    To evaluate our approach with respect to the proposed temporal metrics, we used only two datasets (Funding and Civil Comments Toxicity) since they have a longer temporal window. (The Adult dataset has only 5 time windows at yearly resolution). Results are provided in Table 4 (lower scores are better for each metric; they indicate more fairness and/or more stability). The proposed approach yields a better worst case (lower MB) performance compared to the other baselines in both datasets. Additionally, the proposed approach has lower fluctuation measures with the Toxicity dataset (i.e., lower TS and MBD) but not with the Funding dataset in which the first baseline is performing slightly better (i.e., lower MBD). In all, the proposed approach yields the best performance in 5 of the 6 scenarios (dataset + metric) considered.

    As previously discussed in Section 5.2, in this work the parameters were chosen to reduce $\Delta$ S.P, which provides a reasonable performance in terms of multiple traditional fairness metrics as well as the temporal fairness metrics.

    \begin{table}[h]
    \caption{Results of proposed temporal fairness metrics experimented with two datasets.}
    \begin{tabular}{cccc|ccc}
    \hline
    \multicolumn{1}{l}{} & \multicolumn{3}{c}{\textbf{Funding}}             & \multicolumn{3}{c}{\textbf{Toxicity}}            \\ \hline
    \multicolumn{1}{l}{} & MB ↓            & TS ↓             & MBD ↓           & MB ↓             & TS ↓             & MBD ↓            \\
    \textbf{Vanilla}           & 0.430          & 0.036          & \textbf{0.148} & 0.179          & 0.035          & 0.109          \\
    \textbf{Static}           & 0.261          & 0.034          & 0.158          & 0.155          & 0.037          & 0.107          \\
    \textbf{Dynamic}           & 0.349          & 0.039          & 0.218          & 0.088          & 0.030          & 0.074          \\
    \textbf{Ours}        & \textbf{0.180} & \textbf{0.029} & 0.164          & \textbf{0.072} & \textbf{0.018} & \textbf{0.063} \\ \hline
    \end{tabular}
    \end{table}

\section{Conclusion and Future Work}
    In this paper, we examine the applicability of using an early prevention approach to mitigate bias in advance by experimenting with three different real-world ML applications. We compare our approach to the traditional models that have been widely used in the fairness literature and evaluate the advantages of the anticipatory correction approach. Additionally, we also propose newer fairness metrics that would be suitable when dealing with temporally evolving settings. 
    
    Although we have a used a simple model for future estimation, we are able to see the effect of this approach on bias reduction. The proposed approach yielded best results in terms of most (though not all) metrics across different real world datasets. This trend was consistent across traditional as well as proposed temporal fairness metrics.
    Some degree of variation in results is consistent with past research suggesting the difficulty in reducing different bias metrics simultaneously. A possible approach suggested in the literature is to identify a primary metric for making prioritization depending on the context \cite{lamba2021empirical, makhlouf2020applicability}.
    
    The work described has some limitations. It focuses on a single pre-processing based bias reduction approach and works with a single machine learning approach over three datasets. 
    Yet, by utilizing dynamic learning there are multiple sources of bias that could be involved in the ML pipelines such as uncontrolled data points quality in each time step as well as the model itself. Additionally, since we are learning in a sequential fashion, the distribution of the protected and unprotected group can switch (i.e., what was considered a majority in the past batches might become a minority in the future batch \cite{gomes2019machine}) but we mitigate this issue by applying a re-training model. 
    
    Our future work will focus on investigating the nuances that could lead to this variation in model training by understanding the dynamics for such applications. Besides that, we will utilize a more robust anticipation model and a range of the fairness mitigation methods to understand their applicability in anticipatory bias correction. Our approach is one of the earliest attempts in \textit{anticipatory bias prevention} and we hope that it will encourage the research community to undertake more sophisticated efforts in this direction.

\begin{acks}
This material is in part based upon work supported by the National Science Foundation under Grant SES-1915790.
\end{acks}

\balance
\bibliographystyle{ACM-Reference-Format}
\bibliography{bibliography}

\appendix

\section{Supplementary Material}

\subsection{The effects of $\alpha$ in $\Delta$ S.P}

    To examine the effects of $\alpha$ (i.e., Eq.(3)) on $\Delta$ S.P, Figure 3 shows the range of $\alpha$ on the x-axis and $\Delta$ S.P on the y-axis. Lower values of $\alpha$ focus on future estimates; whereas higher values are emphasising on the current estimates. For the Funding dataset, the $\Delta$ S.P has a decreasing trend as we rely more and more on the current estimates rather than the future estimates in which the lower value is $\alpha = 0.9$ (Fig. 3(a)). Similarly for the Adult dataset, relying on the current estimates has better performance; yet the trend is not smooth (Fig. 3(c)). In contrast, the Toxicity dataset has the opposite effect in which having to rely on the future estimates is much better in order to have a minimal $\Delta$ S.P reduction (Fig. 3 (b)).

\begin{figure*}[h]
  \centering
  \begin{subfigure}[b]{0.4\linewidth}
    \includegraphics[width=\linewidth]{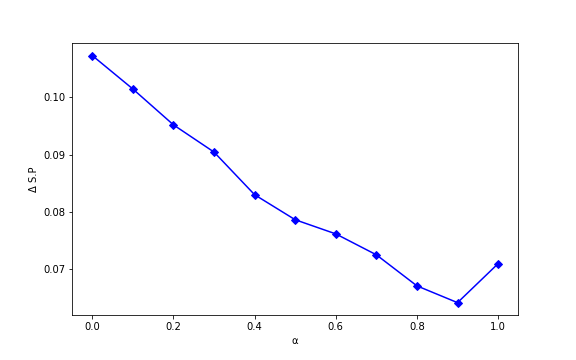}
    \caption{Funding dataset}
  \end{subfigure}
  \begin{subfigure}[b]{0.4\linewidth}
    \includegraphics[width=\linewidth]{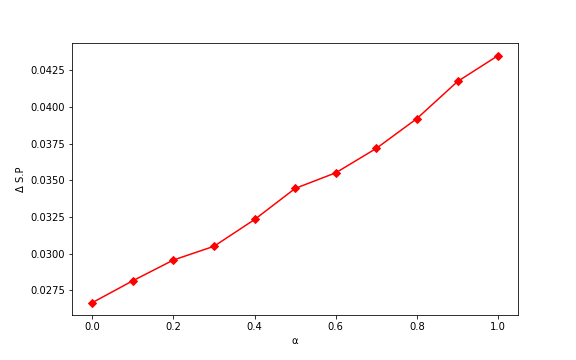}
    \caption{Toxicity dataset}
  \end{subfigure}
    \begin{subfigure}[b]{0.4\linewidth}
    \includegraphics[width=\linewidth]{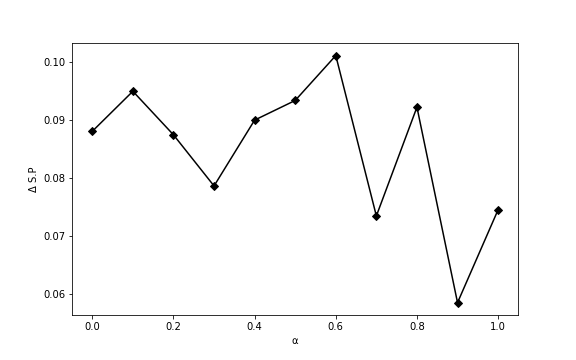}
    \caption{Adult dataset}
    \end{subfigure}
  \caption{Examining the effects of $\alpha$ in $\Delta $ S.P.}
  \label{fig:coffee}
\end{figure*}

\subsection{Temporal Variation}
    In Figure 4, we examine the dynamics of both bias and accuracy through time. We can clearly see that for the Funding dataset that model's accuracy is slightly reducing over time for all models including the baselines. Yet, baseline 2 along with our approach are performing better on average. Similarly for the fairness metrics, our method is able to achieve a bias reduction on average with a small difference between our method and baseline 2. The Toxicity dataset, unlike the Funding dataset, has no obvious trend in terms of the accuracy.  For the fairness metrics, the bias has reduced on average with lesser fluctuations (see $\Delta$ S.P and $\Delta$ TPR) in the later periods. Lastly, the Adult dataset has similar effects on average (except for $\Delta$ FPR) but the small number of temporal points available for analysis makes it less interpretable across time.

    These observations show that bias is rarely stable and constantly changing as long as the data points arrive; thus, a model that is trained once (i.e., typical settings in ML literature) might easily suffer from such issues.

    \begin{figure*}[h]
    \begin{subfigure}{\linewidth}
     \begin{center}
      \includegraphics[width=.70\linewidth]{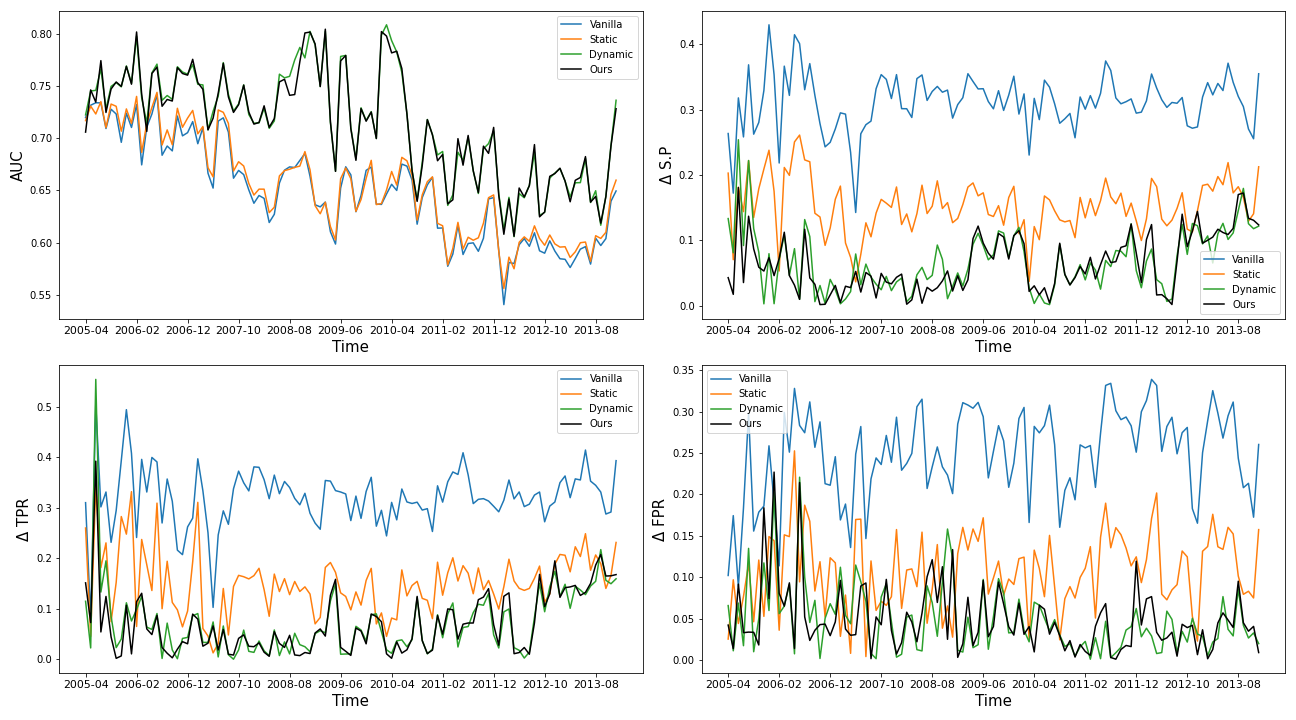}
     \end{center}
      \caption{Funding dataset}
      \end{subfigure}\par\medskip
      
      \begin{subfigure}{\linewidth}
     \begin{center}
      \includegraphics[width=.70\linewidth]{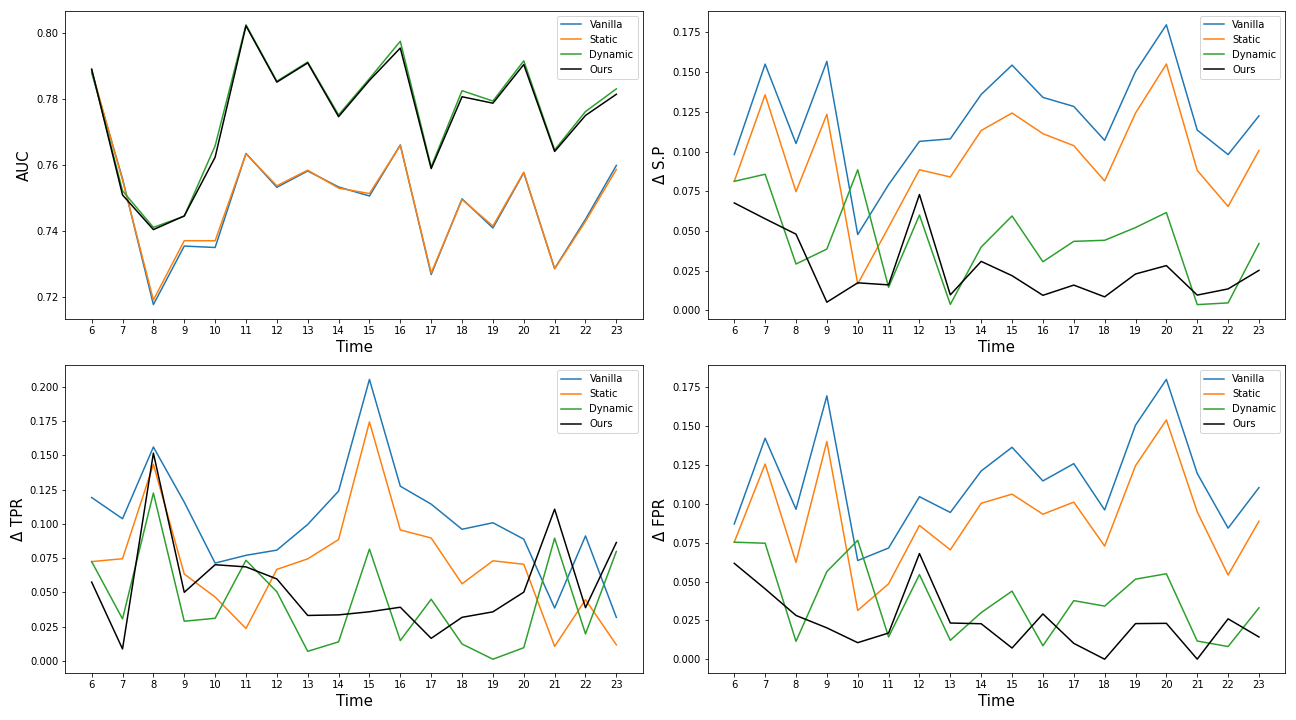}
     \end{center}
      \caption{Toxicity dataset}
      \end{subfigure}\par\medskip
      
      \begin{subfigure}{\linewidth}
     \begin{center}
      \includegraphics[width=.70\linewidth]{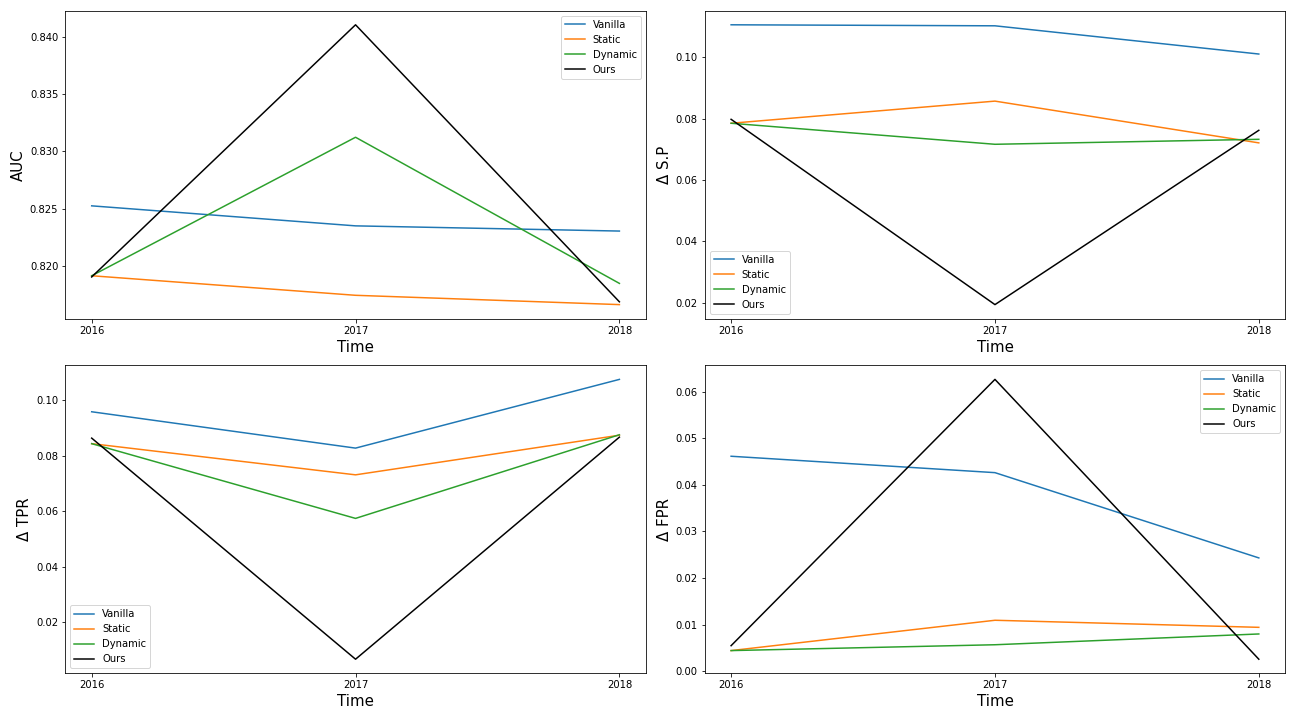}
     \end{center}
      \caption{Adult dataset}
      \end{subfigure}
      \caption{Variations in AUC, $\Delta$ S.P, $\Delta$ TPR, and  $\Delta$ FPR  over time in three different datasets.}
     
    \end{figure*}

\end{document}